\documentclass[11pt]{article}
\usepackage{acl2015}
\usepackage{times}
\usepackage{url}
\usepackage{latexsym}
\usepackage{graphicx}
\usepackage{amsmath}

\title{Exploiting Task-Oriented Resources to Learn Word Embeddings for Clinical Abbreviation Expansion}

\author{Yue Liu$^1$, Tao Ge$^2$, Kusum S. Mathews$^3$,
	Heng Ji$^1$, Deborah L. McGuinness$^1$\\
	$^1$Department of Computer Science, Rensselaer Polytechnic Institute\\
	\{liuy30,jih,dlm\}@rpi.edu\\
	$^2$School of Electronics Engineering and Computer Science, Peking University\\
	getao@pku.edu.cn\\
	$^3$Departments of Medicine and Emergency Medicine, Icahn School of Medicine at Mount Sinai\\
	kusum.mathews@mssm.edu\\
}
  
\date{}

\begin{document}
\maketitle
\begin{abstract}
In the medical domain, identifying and expanding abbreviations in clinical texts is a vital task for both better human and machine understanding. It is a challenging task because many abbreviations are ambiguous especially for intensive care medicine texts, in which phrase abbreviations are frequently used. Besides the fact that there is no universal dictionary of clinical abbreviations and no universal rules for abbreviation writing, such texts are difficult to acquire, expensive to annotate and even sometimes, confusing to domain experts. This paper proposes a novel and effective approach -- exploiting task-oriented resources to learn word embeddings for expanding abbreviations in clinical notes. We achieved 82.27\% accuracy, close to expert human performance.
\end{abstract}

\section{Introduction}

Abbreviations and acronyms appear frequently in the medical domain. Based on a popular online knowledge base, among the 3,096,346 stored abbreviations, 197,787 records are medical abbreviations, ranked first among all ten domains.\footnote{www.allacronyms.com} An abbreviation can have over 100 possible explanations\footnote{www.allacronyms.com/\_medical/HD} even within the medical domain. Medical record documentation, the authors of which are mainly physicians, other health professionals, and domain experts, is usually written under the pressure of time and high workload, requiring notation to be frequently compressed with shorthand jargon and acronyms. This is even more evident within intensive care medicine, where it is crucial that information is expressed in the most efficient manner possible to provide time-sensitive care to critically ill patients, but can result in code-like messages with poor readability. For example, given a sentence written by a physician with specialty training in critical care medicine, ``STAT TTE c/w RVS. AKI - no CTA. .. etc'', it is difficult for non-experts to understand all abbreviations without specific context and/or knowledge. But when a doctor reads this, he/she would know that although ``STAT'' is widely used as the abbreviation of ``statistic'', ``statistics'' and ``statistical'' in most domains, in hospital emergency rooms, it is often used to represent ``immediately''. Within the arena of medical research, abbreviation expansion using a natural language processing system to automatically analyze clinical notes may enable knowledge discovery (e.g., relations between diseases) and has potential to improve communication and quality of care.


\begin{figure}[htbp]
  \centering
      \includegraphics[scale=0.25]{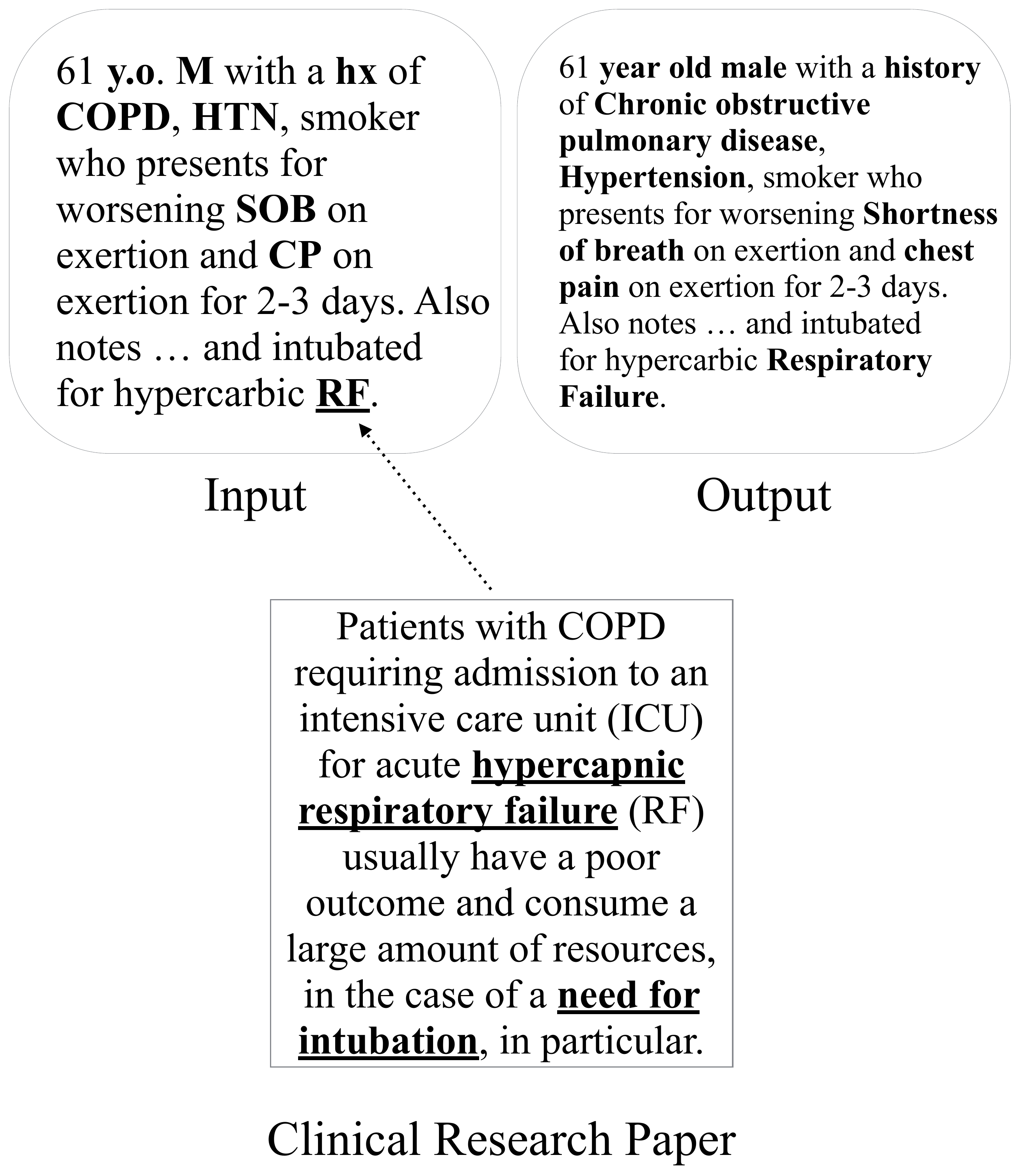}
  \caption{ Sample Input and Output of the task and intuition of distributional similarity}
\end{figure}

In this paper, we study the task of abbreviation expansion in clinical notes. As shown in Figure 1, our goal is to normalize all the abbreviations in the intensive care unit (ICU) documentation to reduce misinterpretation and to make the texts accessible to a wider range of readers. For accurately capturing the semantics of an abbreviation in its context, we adopt word embedding, which can be seen as a distributional semantic representation and has been proven to be effective \cite{mikolov2013distributed} to compute the semantic similarity between words based on the context without any labeled data. The intuition of distributional semantics \cite{harris1954distributional} is that if two words share similar contexts, they should have highly similar semantics. For example, in Figure 1, ``RF'' and ``respiratory failure'' have very similar contexts so that their semantics should be similar. If we know ``respiratory failure'' is a possible candidate expansion of ``RF'' and its semantics is similar to the ``RF'' in the intensive care medicine texts, we can determine that it should be the correct expansion of ``RF''. Due to the limited resource of intensive care medicine texts where full expansions rarely appear, we exploit abundant and easily-accessible task-oriented resources to enrich our dataset for training embeddings. To the best of our knowledge, we are the first to apply word embeddings to this task. Experimental results show that the embeddings trained on the task-oriented corpus are much more useful than those trained on other corpora. By combining the embeddings with domain-specific knowledge, we achieve 82.27\% accuracy, which outperforms baselines and is close to human's performance.


\section{Related Work}

The task of abbreviation disambiguation in biomedical documents has been studied by various researchers using supervised machine learning algorithms \cite{liu2004multi,gaudan2005resolving,yu2006large,ucgun2006predictors,stevenson2009disambiguation}. However, the performance of these supervised methods mainly depends on a large amount of labeled data which is extremely difficult to obtain for our task since intensive care medicine texts are very rare resources in clinical domain due to the high cost of de-identification and annotation. Tengstrand et al. \shortcite{tengstrand2014eacl} proposed a distributional semantics-based approach for abbreviation expansion in Swedish but they focused only on expanding single words and cannot handle multi-word phrases. In contrast, we use word embeddings combined with task-oriented resources and knowledge, which can handle multiword expressions.

\section{Approach}

\subsection{Overview}

\begin{figure}[htbp]
  \centering
      \includegraphics[scale=0.25]{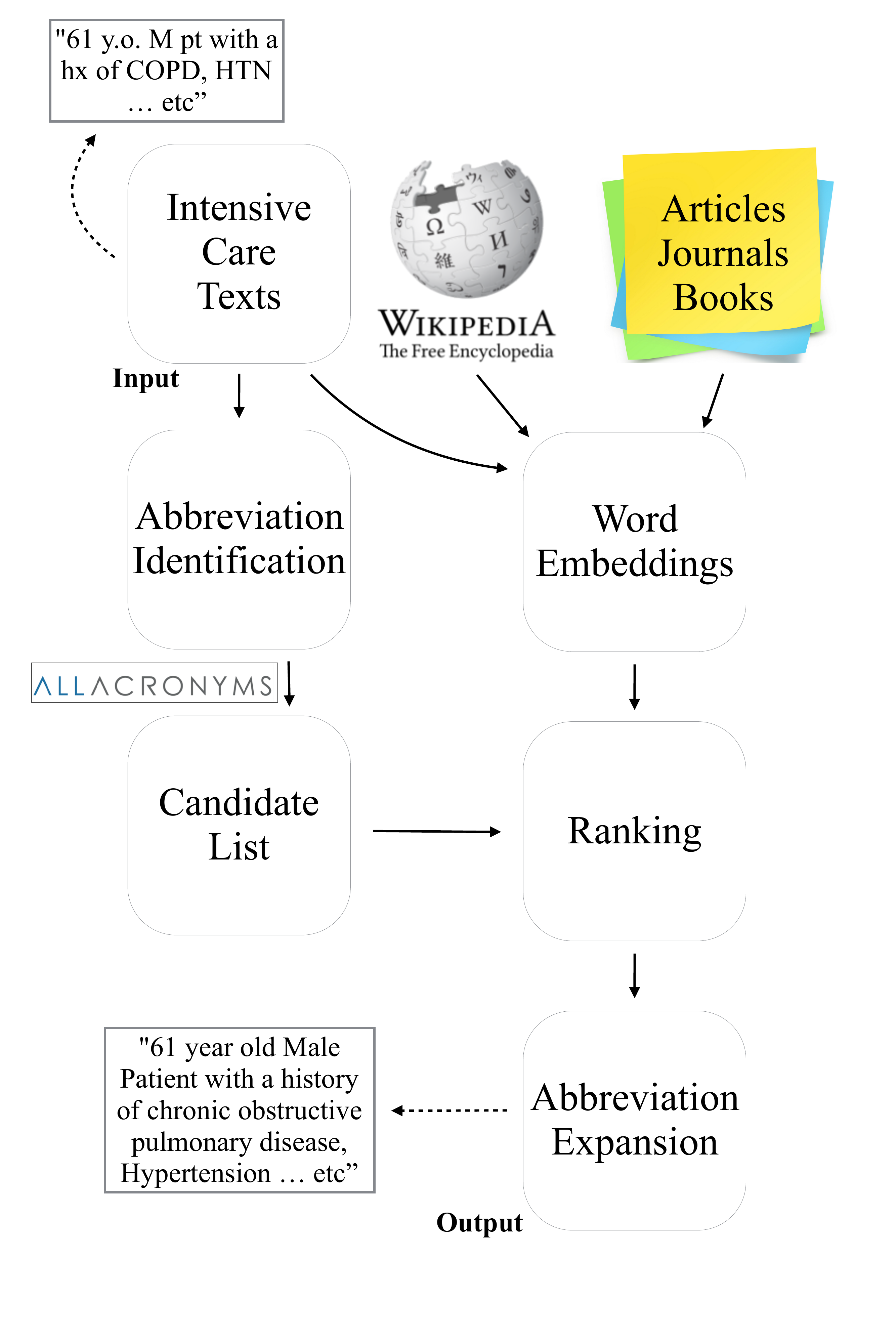}
  \caption{ Approach overview.\label{fig:overview}}
\end{figure}

The overview of our approach is shown in Figure \ref{fig:overview}. Within ICU notes (e.g., text example in top-left box in Figure 2), we first identify all abbreviations using regular expressions and then try to find all possible expansions of these abbreviations from domain-specific knowledge base\footnote{http://www.allacronyms.com} as candidates. We train word embeddings using the clinical notes data with task-oriented resources such as Wikipedia articles of candidates and medical scientific papers and compute the semantic similarity between an abbreviation and its candidate expansions based on their embeddings (vector representations of words).

\subsection{Training embeddings with task oriented resources}

Given an abbreviation as input, we expect the correct expansion to be the most semantically similar to the abbreviation, which requires the abbreviation and the expansion share similar contexts. For this reason, we exploit rich task-oriented resources such as the Wikipedia articles of all the possible candidates, research papers and books written by the intensive care medicine fellows. Together with our clinical notes data which functions as a corpus, we train word embeddings since the expansions of abbreviations in the clinical notes are likely to appear in these resources and also share the similar contexts to the abbreviation's contexts.

\subsection{Handling MultiWord Phrases}
In most cases, an abbreviation's expansion is a multi-word phrase. Therefore, we need to obtain the phrase's embedding so that we can compute its semantic similarity to the abbreviation.

It is proven that a phrase's embedding can be effectively obtained by summing the embeddings of words contained in the phrase \cite{mikolov2013distributed,socher2013reasoning}. For computing a phrase's embedding, we formally define a candidate $c_i$ as a list of the words contained in the candidate, for example: one of \textit{MICU}'s candidate expansions is \textit{medical intensive care unit}=[\textit{medical,intensive,care,unit}]. Then, $c_i$'s embedding can be computed as follows:
\begin{equation}
\boldsymbol{x(c_i)} = \sum_{t \in c_i}\boldsymbol{x(t)}
\end{equation}
where $t$ is a token in the candidate $c_i$ and $\boldsymbol{x(\cdot)}$ denotes the embedding of a word/phrase, which is a vector of real-value entries.

\subsection{Expansion Candidate Ranking}

Even though embeddings are very helpful to compute the semantic similarity between an abbreviation and a candidate expansion, in some cases, context-independent information is also useful to identify the correct expansion. For example, CHF in the clinical notes usually refers to ``congestive heart failure''. By using embedding-based semantic similarity, we can find two possible candidates -- ``congestive heart failure'' (similarity=0.595) and ``chronic heart failure''(similarity=0.621). These two candidates have close semantic similarity score but their popularity scores in the medical domain are quite different -- the former has a rating score\footnote{All the rating information in this paper is from http://www.allacronyms.com. On this website, users are free to rate expansions of an abbreviation if they like the expansions. In general, a popular expansion has a high rating score.} of 50 while the latter only has a rating score of 7. Therefore, we can see that the rating score, which can be seen as a kind of domain-specific knowledge, can also contribute to the candidate ranking.

We combine semantic similarity with rating information. Formally, given an abbreviation $b$'s candidate list $l(b)=\{c_1,c_2,....,c_n\}$, we rank $l(b)$ based on the following formula:

\begin{equation}\label{eq:rerank}
\small
score(c)=\lambda\frac{rating(c)}{\sum_{c_i\in l(b)}rating(c_i)}+(1-\lambda)\frac{\boldsymbol{x(b)} \cdot \boldsymbol{x(c)}}{|\boldsymbol{x(b)}||\boldsymbol{x(c)}|}
\end{equation}
where $rating(c)$ denotes the rating of this candidate as an expansion of the abbreviation $b$, which reflects this candidate's popularity, $\boldsymbol{x(\cdot)}$ denotes the embedding of a word. The parameter $\lambda$ serves to adjust the weights of similarity and popularity\footnote{In the experiments, $\lambda$ is empirically tuned to 0.2 on a separate development set.}

\section{Experiment Results}

\subsection{Data and Evaluation Metrics}

The clinical notes we used for the experiment are provided by domain experts, consisting of 1,160 physician logs of Medical ICU admission requests at a tertiary care center affiliated to Mount Sanai. Prospectively collected over one year, these semi-structured logs contain free-text descriptions of patients' clinical presentations, medical history, and required critical care-level interventions. We identify 818 abbreviations and find 42,506 candidates using domain-specific knowledge (i.e., www.allacronym.com/\_medical). The enriched corpus contains 42,506 Wikipedia articles, each of which corresponds to one candidate, 6 research papers and 2 critical care medicine textbooks, besides our raw ICU data.

We use word2vec \cite{mikolov2013distributed} to train the word embeddings. The dimension of embeddings is empirically set to 100.

Since the goal of our task is to find the correct expansion for an abbreviation, we use \textit{accuracy} as a metric to evaluate the performance of our approach. For ground-truth, we have 100 physician logs which are manually expanded and normalized by one of the authors Dr. Mathews, a well-trained domain expert, and thus we use these 100 physician logs as the test set to evaluate our approach's performance.

\subsection{Baseline Models}

For our task, it's difficult to re-implement the supervised methods as in previous works mentioned since we do not have sufficient training data. And a direct comparison is also impossible because all previous work used different data sets which are not publicly available. Alternatively, we use the following baselines to compare with our approach.

\begin{itemize}
\item Rating: This baseline model chooses the highest rating candidate expansion in the domain specific knowledge base.
\item Raw Input embeddings: We trained word embeddings only from the 1,160 raw ICU texts and we choose the most semantically related candidate as the answer.
\item General embeddings: Different from the Raw Input embeddings baseline, we use the embedding trained from a large biomedical data collection that includes knowledge bases like PubMed and PMC and a Wikipedia dump of biomedical related articles \cite{pyysalo2013distributional} for semantic similarity computation.
\end{itemize}

\subsection{Results}

Table 1 shows the performance of abbreviation expansion. Our approach significantly outperforms the baseline methods and achieves 82.27\% accuracy.

\begin{table}[htbp]
\begin{center}
    \begin{tabular}{|l|c|}
    \hline
   \bf Approaches  & \bf Accuracy \\ \hline
    Rating & 21.32\% \\ \hline
    Raw input embeddings & 26.45\% \\ \hline
    General embeddings & 28.06\% \\ \hline
    Our Approach & \bf 82.27\% \\
    \hline
    \end{tabular}
\end{center}
  \caption{Overall performance}
\end{table}

Figure \ref{fig:rerank} shows how our approach improves the performance of a rating-based approach. By using embeddings, we can learn that the meaning of ``OD'' used in our test cases should be ``overdose'' rather than ``out-of-date'' and this semantic information largely benefits the abbreviation expansion model.

\begin{figure}[h!]
\centering
\begin{itemize}
    \item `OD'- rating-based: [`out-of-date', `other diseases', `on duty', `once daily', `optometry degree', `organ donation', `overdose', `optic disc' ... etc.]
    \item `OD'- our approach: [`overdose', `osteochondritis dissecans', `optic disc' ... ... etc.]
\end{itemize}
  \caption{Ranking lists of expansions of ``OD'' by the rating-based method, our approach\label{fig:rerank}}
\end{figure}

Compared with our approach, embeddings trained only from the ICU texts do not significantly contribute to the performance over the rating baseline. The reason is that the size of data for training the embeddings is so small that many candidate expansions of abbreviations do not appear in the corpus, which results in poor performance. It is notable that general embeddings trained from large biomedical data are not effective for this task because many abbreviations within critical care medicine appear in the biomedical corpus with different senses. 

\begin{figure}[h!]
\centering
\begin{itemize}
    \item Output of general Embeddings on abbreviation `OD': [`O.D.', `optical density', `OD450', `O.D', `OD570', `absorbance', `OD490', `600nm' ... etc.]
\end{itemize}
  \caption{The output of general embeddings trained on large biomedical texts\label{fig:pubmed}}
\end{figure}

For example, ``OD'' in intensive care medicine texts refers to ``overdose'' while in the PubMed corpus it usually refers to ``optical density'', as shown in Figure \ref{fig:pubmed}. Therefore, the embeddings trained from the PubMed corpus do not benefit the expansion of abbreviations in the ICU texts.

Moreover, we estimated human performance for this task, shown in Table \ref{tab:human}. Note that the performance is estimated by one of the authors Dr. Mathews who is a board-certified pulmonologist and critical care medicine specialist based on her experience and the human's performance estimated in Table \ref{tab:human} is under the condition that the participants can not use any other external resources. We can see that our approach can achieve a performance close to domain experts and thus it is promising to tackle this challenge.

\begin{table}[htbp]
\centering
\begin{tabular}{|p{5.3cm}|c|}
\hline
\bf Groups & \bf Accuracy \\ \hline
General readers & $<$40\% \\ \hline
Nurses & 40\% \\ \hline
Mid-level provider (nurse practitioner or physician associate) & 70\% \\ \hline
General practicing physician & 80\% \\ 
\hline
Domain experts with additional training in Emergency Medicine or Critical Care Medicine & \bf $>$90\% \\
\hline
\end{tabular}
\caption{Estimated human performance for abbreviation expansion}
\label{tab:human}
\end{table}

\subsection{Error Analysis}

The distribution of errors is shown in Table \ref{tab:error}. There are mainly three reasons that cause the incorrect expansion. In some cases, some certain abbreviations do not exist in the knowledge base. In this case we would not be able to populate the corresponding candidate list. Secondly, in many cases although we have the correct expansion in the candidate list, it's not ranked as the top one due to the lower semantic similarity because there are not enough samples in the training data. Among all the incorrect expansions in our test set, such kind of errors accounted for about 54\%. One possible solution may be adding more effective data to the embedding training, which means discovering more task-oriented resources. In a few cases, we failed to identify some abbreviations because of their complicated representations. For example, we have the following sentence in the patient's notes: `` No n/v/f/c.'' and the correct expansion should be ``No nausea/vomiting/fever/chills.'' Such abbreviations are by far the most difficult to expand in our task because they do not exist in any knowledge base and usually only occur once in the training data.

\begin{table}[htbp]
\begin{center}
    \begin{tabular}{|l|c|}
    \hline
   \bf Type of error  & \bf Percentage \\ \hline
    Out of Vocabulary & 27\% \\ \hline
    Lack of training samples & 54\% \\ \hline
    Unidentified representation & 19\% \\
    \hline
    \end{tabular}
\end{center}
  \caption{Error distribution}
  \label{tab:error}
\end{table}

\section{Conclusions and Future Work}
\label{sec:length}

This paper proposes a simple but novel approach for automatic expansion of abbreviations. It achieves very good performance without any manually labeled data. Experiments demonstrate that using task-oriented resources to train word embeddings is much more effective than using general or arbitrary corpus.

In the future, we plan to collectively expand semantically related abbreviations co-occurring in a sentence. In addition, we expect to integrate our work into a natural language processing system for processing the clinical notes for discovering knowledge, which will largely benefit the medical research.

\section*{Acknowledgements}

This work is supported by RPI's Tetherless World Constellation, IARPA FUSE Numbers D11PC20154 and
J71493 and DARPA DEFT No. FA8750-13-2-0041. Dr. Mathews' effort is supported by Award \#1K12HL109005-01 from the National Heart, Lung, and Blood Institute (NHLBI). The content is solely the responsibility of the authors and does not necessarily represent the official views of NHLBI, the National Institutes of Health, IARPA, or DARPA.

\bibliographystyle{acl}
\bibliography{acl}

\end{document}